\documentclass[runningheads]{llncs}

\usepackage{eccv}

\usepackage{eccvabbrv}

\usepackage{graphicx}
\usepackage{booktabs}

\usepackage[accsupp]{axessibility}  

\usepackage[pagebackref,breaklinks,colorlinks,citecolor=eccvblue]{hyperref}
\usepackage{hyperref}

\usepackage{orcidlink}

\begin{document}

\title{Region-Adaptive Transform with Segmentation Prior for Image Compression} 


\author{
Yuxi Liu\inst{1,2}\orcidlink{0009-0007-3385-6306} \and
Wenhan Yang\inst{3}\orcidlink{0000-0002-1692-0069} \and
Huihui Bai\inst{1,2}\orcidlink{0000-0002-3879-8957}  \and
Yunchao Wei\inst{1,2,3}\orcidlink{0000-0002-2812-8781} \and \\ 
Yao Zhao\thanks{Corresponding author}\inst{1,2,3}\orcidlink{0009-0008-9174-5433}
}

\authorrunning{Y. Liu et al.}

\institute{Institute of Information Science, Beijing Jiaotong University, Beijing, China \and
Visual Intelligence + X International Joint Laboratory of the Ministry of Education \and
Pengcheng Laboratory, Shenzhen, China \\
\email{\{yuxiliu, hhbai, yunchao.wei, yzhao\}@bjtu.edu.cn}, \quad \email{yangwh@pcl.ac.cn}}

\maketitle

\newcommand{\wh}[1]{{\color{black}#1}}
\newcommand{\wht}[1]{{\color{black}#1}}
\newcommand{\yx}[1]{{\color{black}#1}}
\newcommand{\new}[1]{{\color{black}#1}}

\begin{abstract}
 Learned Image Compression (LIC) has shown remarkable progress in recent years. Existing works commonly employ CNN-based or Transformer-based modules as transform methods for compression. However, there is no prior research on neural transform that focuses on specific regions. 
    In response, we introduce the class-agnostic segmentation masks 
    \yx{(\textit{i.e.} semantic masks without category labels)} 
    for extracting region-\yx{adaptive}
    contextual information.
    Our proposed module, Region-Adaptive Transform, \wh{applies adaptive} convolutions on different regions guided by the masks. 
    Additionally, we introduce a plug-and-play module named Scale Affine Layer to incorporate \wh{rich} contexts from various regions.
    While there have been prior image compression efforts that involve segmentation masks \wh{as additional intermediate inputs}, our approach differs significantly from them.
    \wh{Our advantages lie in that,} to avoid extra bitrate overhead, we treat these masks as privilege information, which is accessible \wh{during the model training stage}  \yx{but not required}
    during the inference phase.
    \wh{To the best of our knowledge,} we are the first to employ class-agnostic masks as privilege information and achieve superior performance in pixel-fidelity metrics, such as Peak Signal to Noise Ratio (PSNR).
    The experimental results demonstrate our improvement compared to previously well-performing methods, with about 8.2\% bitrate saving compared to VTM-17.0.
    The source code is available at \href{https://github.com/GityuxiLiu/SegPIC-for-Image-Compression}{\texttt{https://github.com/GityuxiLiu/SegPIC-for-Image-Compression}}.
    
\end{abstract}

\section{Introduction}
\label{sec:intro}

With the advent of the multimedia era, an ever-growing volume of image data has emerged on the internet.
These vast volumes of image data demand larger transmission bandwidth and storage capacity.
In this context, the importance of image compression technology has become even more pronounced, \wh{evoking greater efforts in} efficient image compression techniques.
\begin{figure}[t]
  \centering
  \includegraphics[width=7cm]{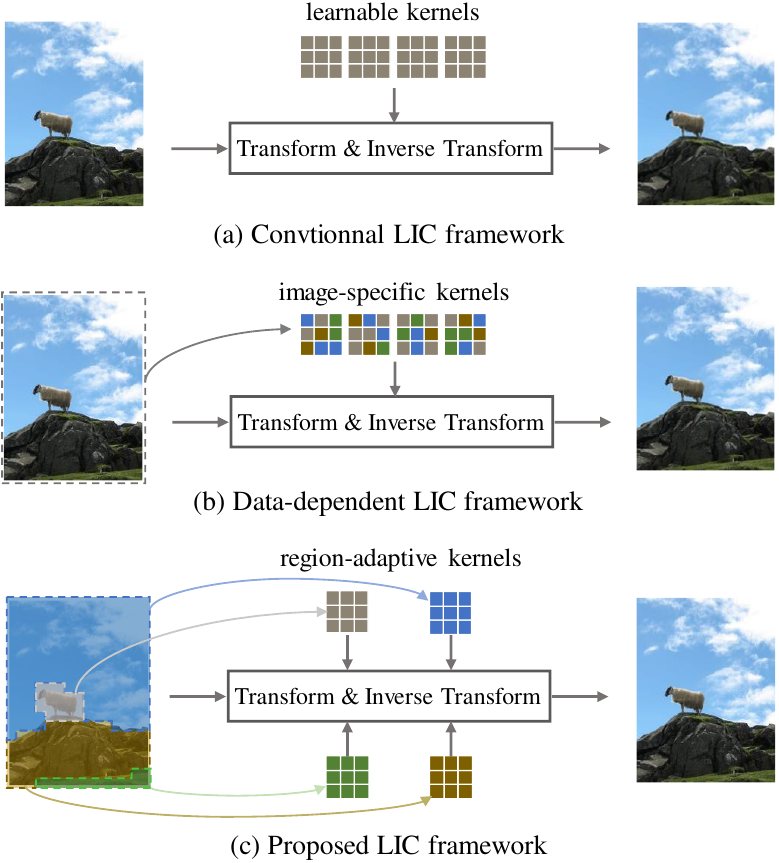}
  \caption{
  \wh{
  Paradigms of end-to-end \yx{learned image compression (LIC) methods.} 
  (a) Conventional LIC with fixed learned \yx{kernels}. (b) \yx{Data}-dependent LIC with image-specific kernels \cite{wang2022neural}. 
  (c) The proposed method is capable of generating region-adaptive kernels from different regions, providing more fine-grained transforms.
  }
  }
  \label{sum}
\end{figure}
Lossy image compression technology can reduce the data size to a lower level while maintaining good visual quality. 
In the past decades, the classical lossy image compression methods such as JPEG~\cite{jpeg}, JPEG2000~\cite{jpeg2000}, BPG~\cite{bpg}, and VVC~\cite{vvc}, 
used to have excellent performance and were widely adopted.
These methods include modules of transform, quantization, and entropy coding.
The transform process \wh{aims to convert} images into a more compact \wh{coefficient} by \wh{removing} pixel correlations \wh{as much as possible}. 
\wh{The commonly employed transforms} include the discrete cosine transform (DCT)~\cite{dct} and wavelet transform~\cite{wavelet}.
\wh{These transformations \yx{are linear}, characterized by simplicity and efficiency.
However, \yx{their} fixed forms are unable to adapt to the contexts of complex signals.}
\wh{To address this issue, \wh{latter efforts propose} data-driven transformation approaches, \textit{e.g.} the Karhunen-Loéve transform (KLT)~\cite{klt}.
}
\yx{
Furthermore, the subsequent works are dedicated to constructing complex patterns for the adaptive selection of local transforms, improving adaptability and overall performance.} 

\wh{Benifit from the advancement of neural network technology, many learned image compression (LIC) methods have been proposed, which shows a promising alternative for designing transform: learning from data instead of handcraft design.}
Balle~\textit{et al.}~\cite{balle2016,balle2018} proposed the first well-known end-to-end framework.
They utilize variational auto-encoders (VAE)~\cite{vae} as a transform module and 
optimize the parameters of the kernels to better align with the natural data distribution.
To further \wh{improve} the representation capabilities of the transform module, 
some works~\cite{cheng2020learned,liu2020unified,stf2022,zhu2021transformer,wang2022neural} 
\wh{design advanced modules to introduce richer contextual information,}
\textit{e.g.} residual blocks~\cite{he2016deep} and self-attention mechanism~\cite{vit,liu2021swin}.
These works optimize learnable kernels to fit the data distribution to attain a more powerful transform. 
However, their kernel parameters are fixed after training (see Fig.\ref{sum} (a)).
Wang \textit{et al.}~\cite{wang2022neural} introduce a neural data-dependent transform,
which extracts dynamic convolution kernels \wh{based on the context of the entire image} for \wh{more} adaptive transform (see Fig.~\ref{sum} (b)).
\wh{However, even within a single image, the context could vary significantly across different regions, 
\textit{e.g.} stone and sheep regions in Fig.~\ref{sum}.
It limits the description of the signals to a more fine-grained level using such image-level transforms.}

To address the issue, this paper
makes the first efforts 
to design learnable region-adaptive transform to capture the mapping regularity in different regions.
In detail, we employ class-agnostic\footnote{``class-agnostic'' means distinguishing different objects in an image regardless of what those objects are. Such masks can be regarded as a generalized representation of partition maps widely used in conventional codecs.} segmentation masks to extract region-specific information for transform (see Fig.~\ref{sum} (c)).
\yx{Differs from previous works~\cite{chang2021thousand,akbari2019dsslic,hoang2020image,feng2023semantically}, our approach stands out with two characteristics in mask utilization.}
\wh{First, our masks are ``class-agnostic'', where the \yx{semantic} labels are abandoned but a more effective compression-driven semantic prior will be
\yx{learned by our model in a data-driven manner.}
}
\yx{Second, to avoid incurring additional costs in both bitrate and computational resources, we treat these masks as \textit{privilege information}, which are available during training but remain inaccessible during the inference phase~\cite{privilege}.
}
Specifically, we replace the masks with uniform grid partitions~(see Fig.~\ref{gradient}~(c)) in inference, and the experimental result shows approximate performance.
It implies that our model \wh{has learned useful semantic knowledge from} the masks in training,
\yx{incorporating this knowledge into the model (\textit{i.e.} the network's parameters).}
Hence, it remains capable of mining contextual information even from
relatively coarse partition regions, \wh{\textit{i.e.}}, 
the \yx{given} uniform grid partitions.
\yx{This simple operation can effectively maintain the consistency of our model's architecture and parameters, whether or not masks are used at different phases.}
\yx{The more details about the difference between our method and previous semantic-mask-based works will be explained at Sec.~\ref{sec:ralted}.}

Based on the above \wh{motivations}, we propose a novel image compression framework called \textbf{Seg}mentation-\textbf{P}rior-Guided \textbf{I}mage \textbf{C}ompression (SegPIC), equipped with two \yx{key} modules: Region-Adaptive Transform (RAT) and Scale Affine Layer (SAL).
\yx{The RAT adopts region-adaptive kernels for transform, performing the convolution whose weights are generated from the prototypes and the context of latent features under the guidance of masks. The SAL affines the latent features for better mining of semantic contexts in the Encoder and Decoder.}
Experimental results show that SegPIC is superior to previously well-performing image compression methods.
Our main contributions are as follows:
\begin{itemize}
    \item We propose a novel image compression framework called Segmentation-Prior-Guided Image Compression (SegPIC) \wh{to generate more effective region-adaptive transforms in a data-driven manner.}
    \item \wh{Our SegPIC is equipped with two proposed modules: Region-Adaptive Transform (RAT) and Scale Affine Layer (SAL).}
    \yx{The RAT adopts region-adaptive kernels for transform, while the SAL affines the latent features for better mining of semantic contexts.}
    \item We utilize the class-agnostic masks as privilege information to assist the training of SegPIC. 
    \yx{The ``class-agnostic'' property promotes the model to learn a compression-driven semantic prior, while the ``privilege'' way avoids additional bitrate and complexity for masks in inference.}
    \item Experimental results demonstrate that our SegPIC is superior to previously well-performing image compression methods. 
    We also present sufficient ablation experiments to confirm the necessity of the privilege information.
\end{itemize}

\section{Related Works}
\label{sec:ralted}

\noindent \textbf{Learned Image Compression.}
Image compression can be divided into three main components: transform, quantization, and entropy coding. In brief, the transformation module maps the image to a latent feature space. Subsequently, quantization is performed after this phase, resulting in some quantization distortions but achieving lower bitrate. The entropy encoding module estimates the data distribution of the latent features and inputs it to the Arithmetic Coding (AC) module, which generates the bitstream for transmission.

To enhance transform module, some works~\cite{liu2020unified,stf2022,zhu2021transformer,wang2022neural} adopt more powerful modules like self-attention mechanism~\cite{vaswani2017attention,liu2021swin} to explore broader context. 
\cite{liu2023learned} proposed a transform block that mixes CNN and Transformer to combine their advantages.
As for the entropy estimation module, the studies~\cite{minnen2018joint,minnen2020channel,cheng2020learned,he2021checkerboard} usually focus on exploring novel auto-regressive methods to split the latent features into several portions to be decoded in order.  The portions having been decoded will assist in decoding the next portion, through the spatial or channel correlation of these portions. 
This auto-regressive mechanism may introduce additional latency but achieve better performance than Hyper Prior~\cite{balle2018}.
Recently, \cite{jiang2023mlic} integrates different types of correlations present in latent representation, \textit{e.g.} channel-wise, local spatial, and global spatial correlations. Their proposed novel auto-regressive entropy module achieve powerful performance.

Thanks to these great works, the performance of LIC has become so strong that achieving
further enhancements becomes progressively challenging. Therefore, some new directions in compression have emerged, which is also meaningful on certain occasions.  For example, some semantic-based compression methods~\cite{chang19Layered,chang2021thousand} adopt GAN~\cite{gan} to compress images at extremely low bitrate. These methods maintain a good human visual perception effect but usually, poor
pixel-fidelity metrics, \textit{e.g.} Peak Signal to Noise Ratio (PSNR) and Multi-Scale Structural Similarity Index (MS-SSIM)~\cite{msssim}. Specifically, these methods could introduce some synthetic artifacts with low fidelity.
For example, a semantic compression method could reconstruct an image of a bird with different breeds from the original image because the training category labels do not distinguish the bird's breed.
Another direction is compression for machines, which means compressing images for downstream tasks~\cite{feng2023semantically,sun2020semantic}. These methods focus on balancing the bitrate and performance of downstream tasks, not just the pixel-fidelity metrics.

Although some works~\cite{chang19Layered,chang2021thousand,feng2023semantically} about the two directions also adopt masks, our usage is significantly different. The main differences are as follows:
\wh{
1) We utilize the masks as privilege information, without the mask information during the inference phase, while usually, the masks are indispensable for other methods.
2) We employ class-agnostic masks to guide the model to extract rich compression-friendly contexts among semantically related regions, regardless of the category
\yx{that might degrade the fidelity as mentioned above.}
3) The proposed model is optimized towards better pixel-fidelity metrics, \textit{e.g.} PSNR and MS-SSIM.
}

\begin{figure*}[t]
  \centering
  \includegraphics[width=10.5cm]{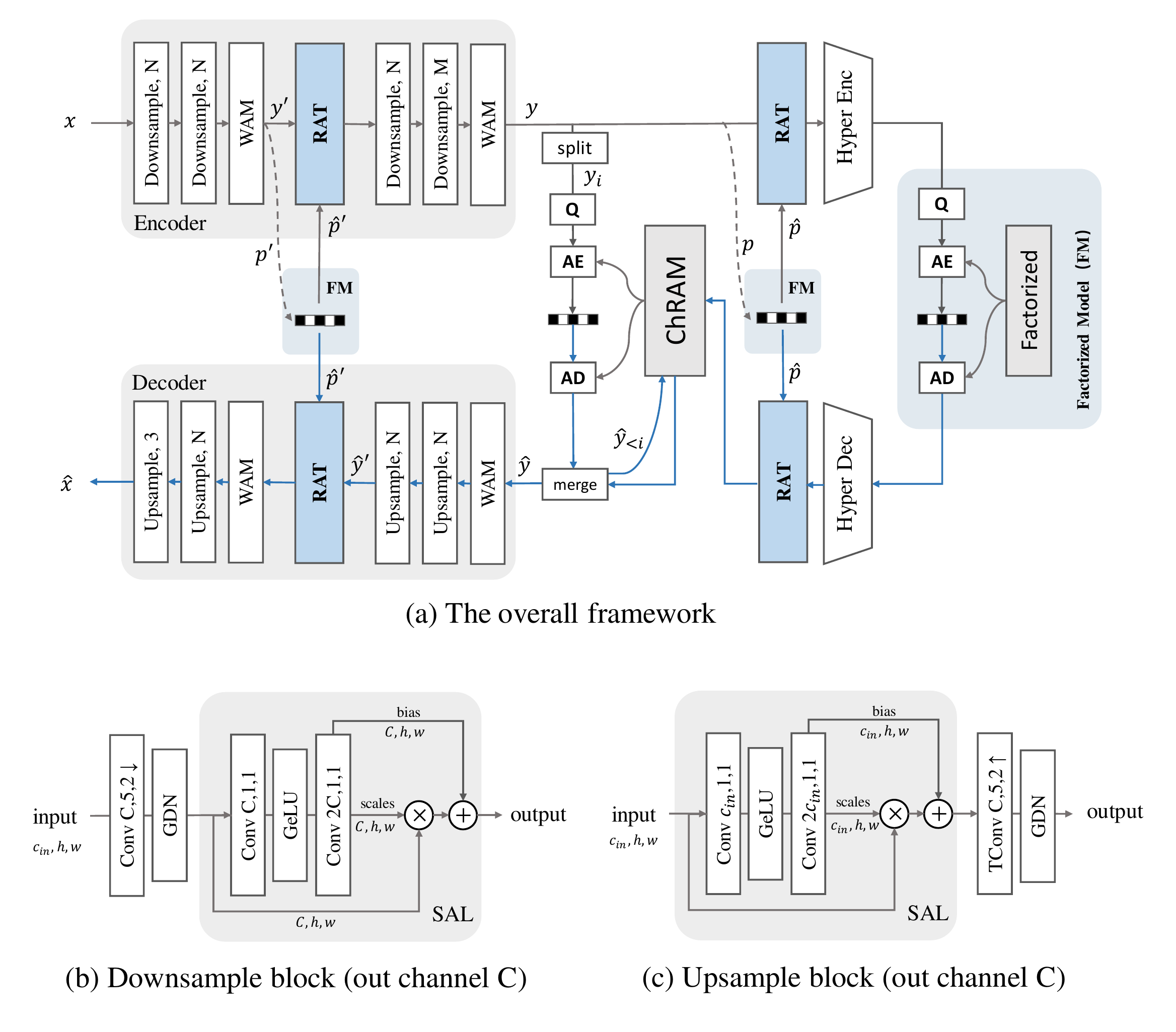}
  \caption{
  (a) \wh{Overall framework of our SegPIC.} 
  The blue flow lines represent the decoding process \wh{on} the decoder side. 
  The dashed lines represent the process of extracting the prototypes $p',p$ (see Eq.~\ref{eq.map}). The compression of $p',p$ is not fully depicted for simplicity, please refer to Fig.~\ref{pro_compress} for details. 
  $N$ is set to 192, and $M$ is 320 as in \cite{stf2022}. 
  RAT is the proposed Region-Adaptive Transform. SAL is the proposed Scale Affine Layer. WAM is Window Attention Module~\cite{stf2022}. 
  ChARM is Channel-wise Auto-Regressive Model~\cite{minnen2020channel}. FM is a Factorized Model~\cite{balle2018}. 
  GDN is Generalized \wh{Divisive Normalization}~\cite{balle2016}. 
  (b) is Downsample Block. ``Downsample, N'' in (a) means the block with out-channel N. ``Conv C,5,2'' means out-channel C, kernel size 5, stride 2. 
  (c) is Upsample Block, and ``TConv'' is Transposed Convolution. }
  \label{arch}
\end{figure*}

\new{
\noindent \textbf{Region-adaptive Convolution.}
Due to its \wht{capacity} to model the degradation levels in different regions of images, region-adaptive convolution is applied to some low-level vision tasks such as super-resolution and denoising~\cite{shen2023adaptive,xu2020unified}.
In terms of compression, the reconstruction quality varies across different regions. For instance, regions with complex textures are more challenging compared to those with simpler \wht{patterns}.
Therefore, specific processing can be applied to different regions based on their characteristics.
The regions can be divided using segmentation masks or \wht{partitions based on certain regular patterns}.
To the best of our knowledge, we are the first to introduce this mechanism on image compression for optimizing pixel-fidelity metrics.
}

\section{Proposed SegPIC}
\subsection{Motivation}

\wh{As mentioned above, previous transforms either adopt convolutions with fixed parameters or image-specific convolutions.
Our motivation is to construct a region-adaptive transform via the following three properties:
\begin{itemize}
    \item \textbf{Region-dependent}: we propose a neural network with region-adaptive transforms guided by the segmentation masks.
    \item \textbf{Class-Agnostic}: the adopted mask does not include a \yx{category} label, 
    \yx{allowing the model to be free to learn compression-friendly semantic priors.}
    \item \textbf{Privilege Information}: the masks are expected as privilege information, \yx{not being used} in inference for saving bitrate and computation resources.
\end{itemize}}

Based on these considerations, we propose our SegPIC in the VAE framework incorporated with two proposed modules: SAL and RAT.

\begin{itemize}
    \item \textbf{SAL}:
    First of all, we extract a prototype from each region, which will be compressed and transmitted to be accessible on both the encoder and decoder sides.
    \new{
    The prototypes are expected to extract high-level information from regions to ensure the representative ability as much as possible. However, the encoder in compression frameworks often extracts \wht{features that contain low-level information}. Therefore, \wht{a module is highly desirable for communicating information across different levels}.
    SAL is designed for extracting more abundant semantics and contexts, enriching the prototypes, and connecting features from two different levels.
    }
    
    \item \textbf{RAT}: 
    performing the
    \yx{Depth\&Point-wise Separable Convolution (DPSConv), a novel convolution whose weights are generated from the prototypes
    and the context of latent features under the guidance of masks.}
    Note that, we only utilize the masks to guide RAT to learn more compression-friendly semantic priors in training, 
    while during the inference phase, we instead adopt uniform grid partitions as coarse alternatives.
    
\end{itemize}

\subsection{Formulation}
In this section, we will introduce the problem and our method in mathematical form. Given an image $x$, the encoder $En$ maps it to latent features $y$. After quantization $Q$, $\hat{y}$ will be compressed to the bitstream by the entropy coding, such as arithmetic coding (AC). Then the receiver converts the bitstream back into $\hat{y}$ by entropy decoding. The above process can be formulated as follows:
\begin{equation}
    \begin{aligned}
        &y=En(x), \quad  \hat{y} = Q(y), \\
        &\text{Transmiter: } \quad bit_{\hat{y}} = AE(\hat{y}|p_{\hat{y}}), \\
        &\text{Receiver: } \quad \quad \hat{y} = AD(bit_{\hat{y}}|p_{\hat{y}}),\\
    \end{aligned}
\end{equation}
where $ p_{\hat{y}}$ is the estimated distribution of $\hat{y}$, $bit_{\hat{y}}$ is the binary bitstream of $\hat{y}$, \wh{and} $AE$ and $AD$ are arithmetic encoding and decoding. We use a single Gaussian distribution to estimate $\hat{y} \sim \mathcal{N}(\mu,\sigma^2)$, \wh{and} the mean and variance value ${\mu,\sigma^2}$ will be yielded through an auto-regressive mechanism. The formulation is:
\begin{equation}
    \begin{aligned}
        &z=En_z(y),\hat{z}=Q(z),    \quad 
        \mu_i,{\sigma_i} = F(De_z(\hat{z}),\hat{y}_{<i}), \\
    \end{aligned}
    \label{eq.z}
\end{equation}

where $En_z, De_z$ is hyper encoder and decoder, $\mu_i,\sigma_i$ are the mean and variance of $\hat{y}_i\sim \mathcal{N}(\mu_i,{\sigma_i}^2)$, $\hat{y}_i$ is the currently decoded portion of $\hat{y}$, $\hat{y}_{<i}$ is the already decoded portion to assist decoding $\hat{y}_i$, and $F$ is the auto-regressive entropy model.
The AC process of $z$ in Eq.~\ref{eq.z} is omitted for brevity. 
Finally, the receiver adopts decoder $De$ to reconstruct the image $\hat{x}$, which is formulated as $\hat{x} = De(\hat{y})$.

The loss function $\mathcal{L}$ to optimize this problem is:
\begin{equation} \label{eq.loss}
    \mathcal{L} = R + \lambda \cdot D = R(\hat{y}) + R(\hat{z})+ \lambda \cdot \mathbb{E}_{x \sim p_x} \left[ d(x,\hat{x}) \right],
\end{equation}
where $R(\hat{y}), R(\hat{z})$ are the bitrate of $\hat{y},\hat{z}$,
$d(x,\hat{x})$ is the distortion between $x$ and $\hat{x}$, $\lambda$ denotes different rate-distortion trade-offs. Here we use Mean Square Error (MSE) or (1-MS-SSIM) as the distortion function.

In our implementation, we use class-agnostic semantic masks~(denoted as $m$) as additional inputs of the encoding $En$ during the training phase. Moreover, we extract a prototype from each region at the middle layer of $En$, then compress and transmit them (denoted as $\hat{p}$). In inference, we replace the masks with uniform grid partitions, so there is no additional bitstream for masks, while the bitrates of the prototypes also need to be taken into account. The $En$ and $De$ will transform the image with the assistance of the prototypes.
The formulations requiring adjustment are as follows:
\begin{equation} \label{eq.ours}
\begin{aligned}
    &y=En(x,\hat{p} | m), \quad \hat{x} = De(\hat{y}, \hat{p}|m),\\
    &\mu_i,\sigma_i = F(De_z(\hat{z}),\hat{y}_{<i}, \hat{p}| m), \\
    &R = R(\hat{p}) + R(\hat{y}) + R(\hat{z}). \\
\end{aligned}
\end{equation}

\subsection{Network Design}
In this section, we will introduce the concrete process and our network design. We use \cite{stf2022} as the backbone. The overall architecture is in Fig.~\ref{arch}. 

\noindent \textbf{Scale Affine Layer.} 
The Encoder and Decoder are responsible for image transform and reconstruction, each comprising 4 downsampling or upsampling blocks.
The previous sampling blocks consisted of a single sample convolution layer, aimed at modifying the data size. However, this may not be sufficient to extract adequate semantics and contexts.
Inspired by Spatial Feature Transform (SFT) \cite{18sft}, 
\new{
which can generate visually pleasing textures in segmentation-guided low-level vision,
}
we propose to add a Scale Affine Layer (SAL) between two adjacent convolution layers (see Fig.~\ref{arch} (b) (c) ). 
It consists of two 1×1 convolution layers with a GELU activation function in between, which is formulated as:
\begin{equation}
\begin{aligned}
       &X_s = \textup{SAL}_s(X),\quad X_b = \textup{SAL}_b(X),\\
       &Y = X_s\otimes X + X_b, \\
\end{aligned}
\nonumber
\end{equation}
where $X$ and $Y$ represent the input and output, $X_s$ and $X_b$ denote the scales and biases generated by SAL,
$\otimes$ denotes the element-wise product.

\begin{figure}[t]
  \centering
  \includegraphics[width=\linewidth]{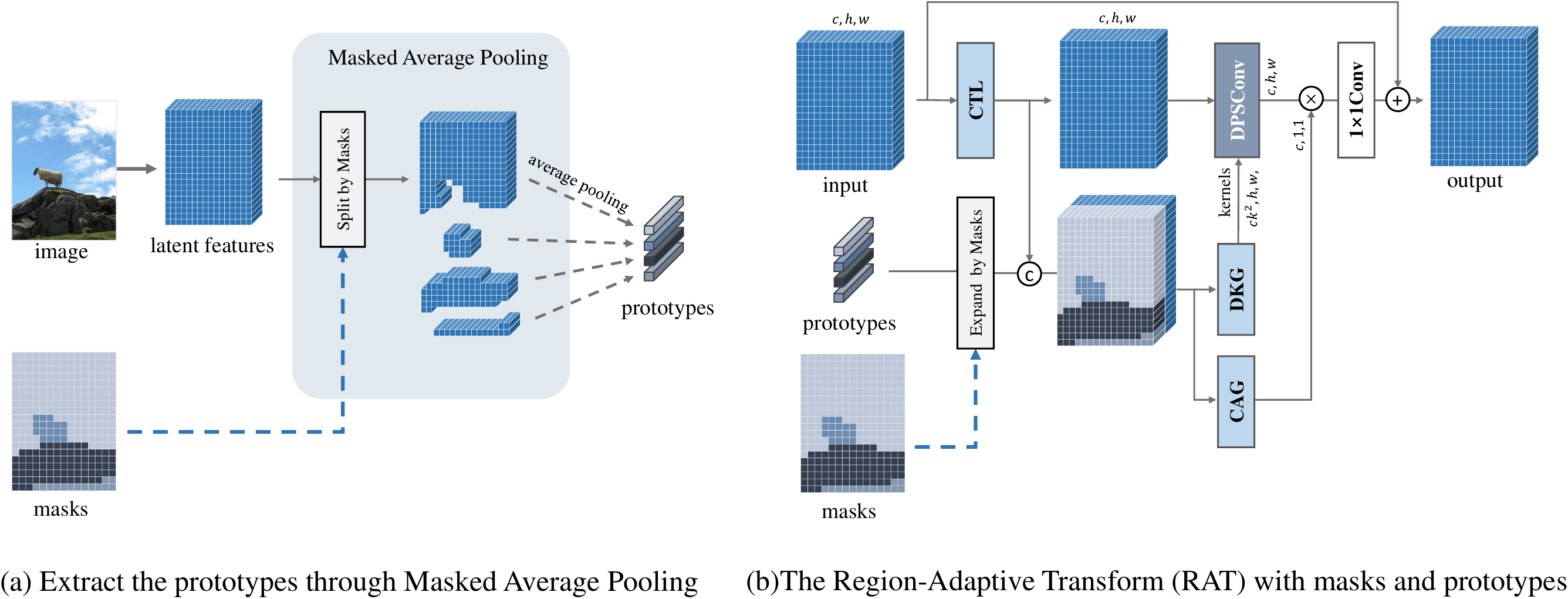}
  \caption{The diagram of extracting the prototypes and the proposed RAT module.
  DPSConv is proposed Depth\&Point-wise Separable Convolution (see Eq.~\ref{eq.dpsconv}). 
  CTL is Channel Transform Layers, DKG is the DPSConv Kernel Generator, and CAG is the Channel Attention Generator. DPSConv is Depth\&Point-wise Separable Convolution. 
  The detailed modules are represented in Fig.~\ref{modules}.
  }
  \label{map}
\end{figure}

\begin{figure}[t]
    \begin{minipage}[b]{0.4\textwidth}
        \centering
        \begin{subfigure}[b]{\textwidth}
            \centering
            \includegraphics[width=\textwidth]{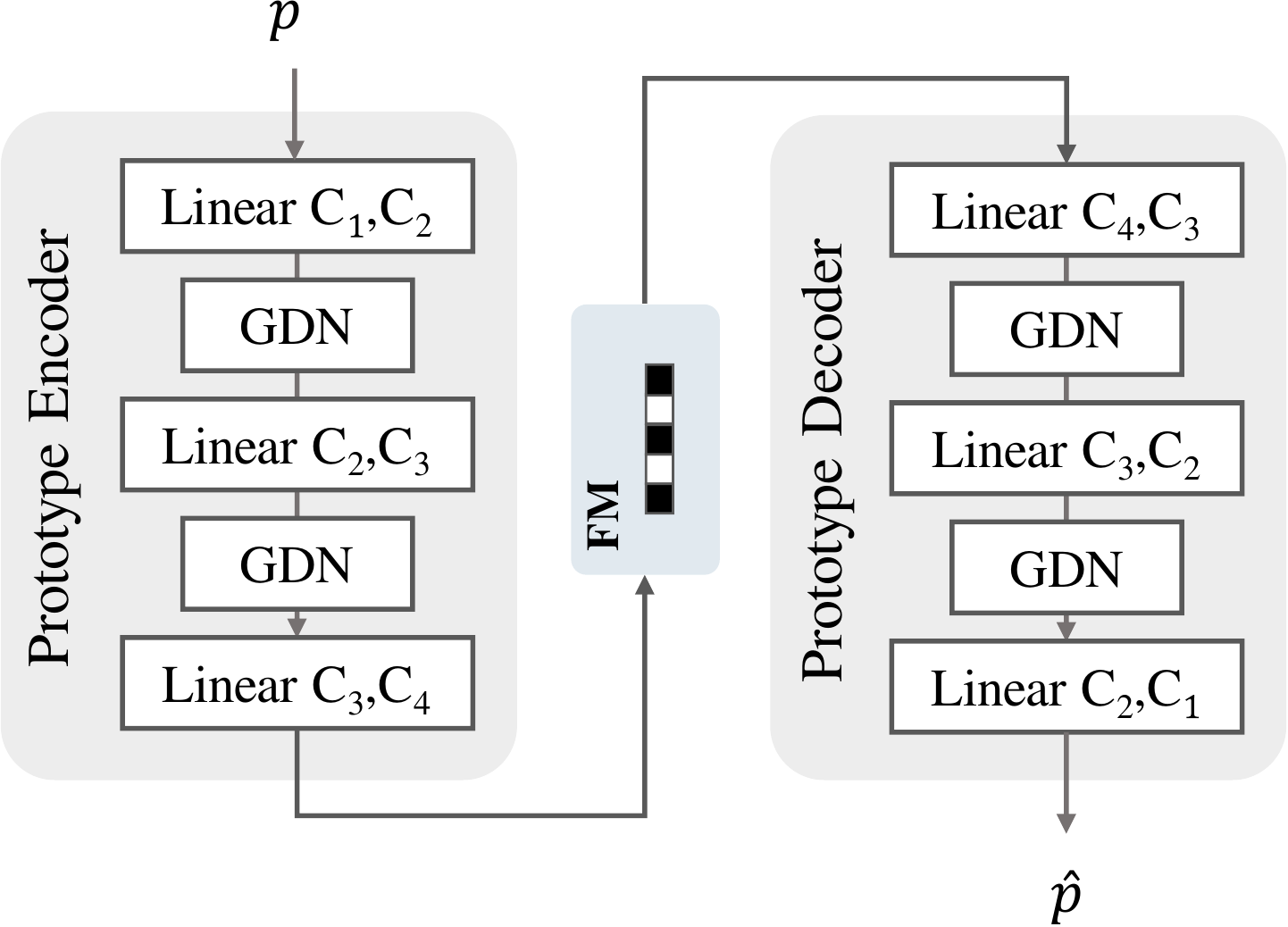}
        \end{subfigure}
        \caption{The architecture of the Prototype Encoder and Decoder for compressing and transmitting the prototypes. 
  ``Linear $\text{C}_1$,$\text{C}_2$'' means linear layer with in-channel $\text{C}_1$ and out-channel $\text{C}_2$.}
        \label{pro_compress}
    \end{minipage}
    \hspace{0.3cm}
    \begin{minipage}[b]{0.57\textwidth}
        \centering
        \begin{subfigure}[b]{\textwidth}
            \centering
            \includegraphics[width=0.8\textwidth]{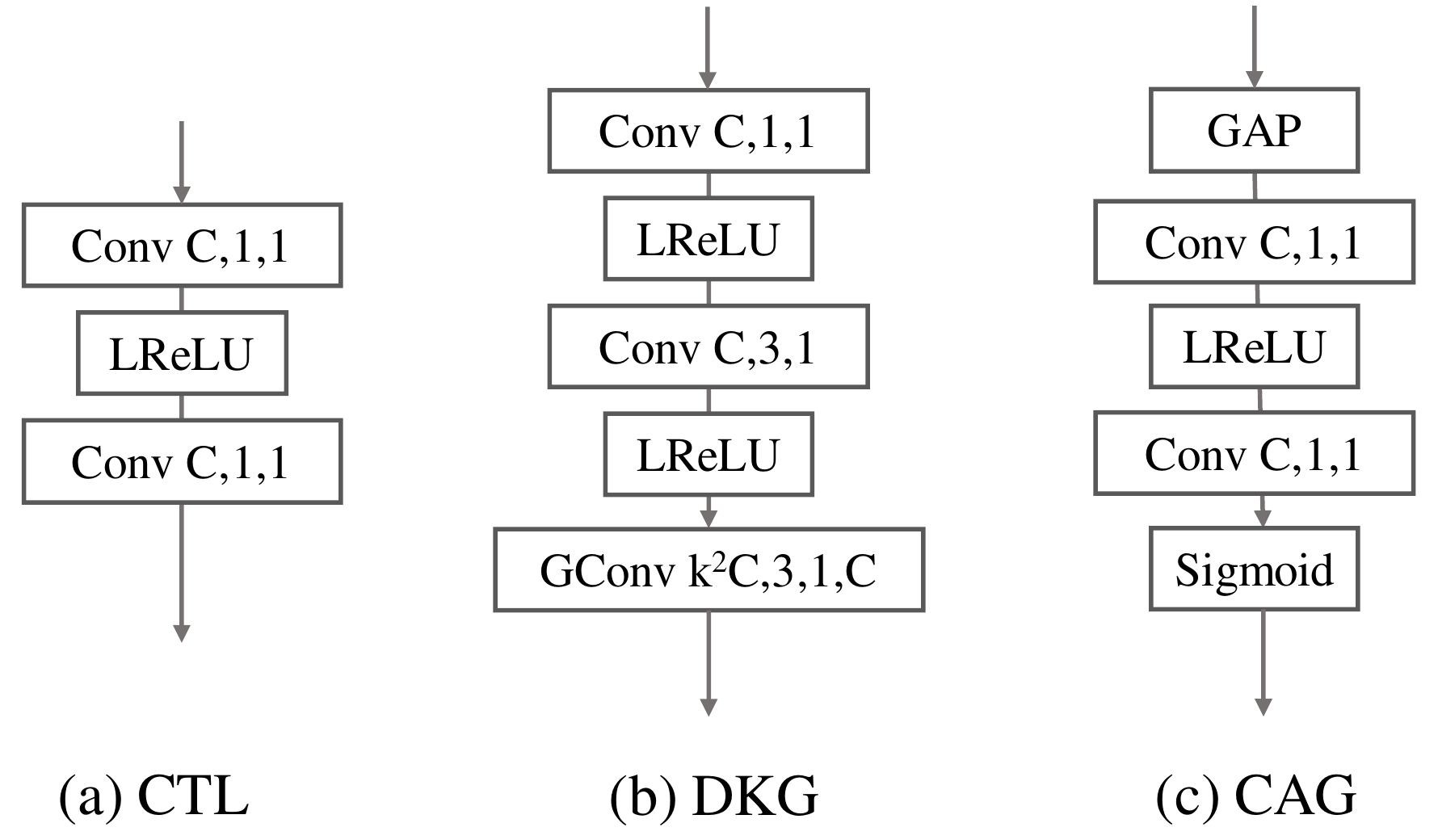}
        \end{subfigure}
        \caption{The detailed modules in RAT, including Channel Transform layers (CTL), DPSConv Kernels Generator (DKG), and Channel Attention Generator (CAG). 
  LReLU is Leaky ReLU. GAP is Global Average Pooling. 
  ``GConv $\text{k}^2$C,3,1,$C$'' means Grouped Convolution with out-channel $\text{k}^2$C, kernel size 3, stride 1 and group $C$.}
        \label{modules}
    \end{minipage}
\end{figure}

\noindent \textbf{Extract and Transmit the Prototypes.} 
The Encoder first converts the original image $x \in \mathbb{R}^{3 \times H \times W}$ into latent features $ y' \in \mathbb{R}^{N \times H/4 \times W/4} $ at the middle layer. Then we conduct Masked Average Pooling (MAP) on $y'$ through the class-agnostic masks $m \in \{ 0,1 \} ^{n \times H \times W}$ (see Fig.~\ref{map}~(a)), to acquire the prototypes $p' \in \mathbb{R}^{N \times n}$, where $n$ is the number of the masks. Thanks to the proposed SAL, the Encoder becomes deeper, allowing $p'$ to incorporate more semantics and contexts from the corresponding mask. The formulation of MAP is as follows: 
\begin{equation}
       p' = \textup{MAP}(y', m) = \{\frac{1}{\sum m^{(i)}}\sum y' \otimes m^{(i)}, ... \}_{i \in \left[1,n\right]},
\label{eq.map}
\end{equation}
where $y'$ are the mid-level features (see Fig.~\ref{arch}~(a)), $\sum$ denotes summation along the spatial dimension.
$p'$ will be compressed and transmitted to the receiver (see Fig.~\ref{pro_compress}), and the reconstructed $\hat{p}'$ is shared between the Encoder and Decoder. The channel \{$C_1, C_2, C_3, C_4$\} is \{192, 128, 96, 96\} for $p'$ (on Encoder and Decoder) and  \{320, 192, 128, 96\} for $p$ (on Entropy Model). 

\noindent \textbf{Region-Adaptive Transform.}
Here we propose a novel convolution called Depth\&Point-wise Separable Convolution (DPSConv), in which each element has an individual convolution kernel, 
and there is no channel-wise mutual communication for computational efficiency.

Before introducing it, we will start with standard convolution briefly. 
The input of standard convolution can be denoted as $X \in \mathbb{R}^{C\times H \times W}$, 
the output is $Y \in \mathbb{R}^{O\times H \times W}$, and the kernel is $W \in \mathbb{R}^{O\times C \times k \times k}$, where $k$ is kernel size.
The formulation of standard convolution is as follows:
\begin{equation}
       Y_{o,h,w} = \sum\limits_{c=1}^C X_{c,h,w}\ast W_{o,c}=\sum\limits_{c=1}^C \sum\limits_{i,j} X_{c,h+i,w+j} \cdot W_{o,c,i,j},
       \nonumber
\end{equation} 
where $\ast$ is 2D convolution,  
$i,j\in [-\lfloor k/2 \rfloor, \lfloor k/2 \rfloor]$, 
$k$ is kernel size, \wh{and} $\lfloor \cdot \rfloor$ denotes floor rounding.

DPSConv can be seen as a combination of Depthwise Separable Convolution (DWConv) \cite{chollet2017xception} and Dynamic Convolution \cite{jia2016dynamic}.
The formulation is as follows:
\begin{equation}
    Y_{c,h,w} = X_{c,h,w}\ast W_{c,h,w} = \sum\limits_{i,j} X_{c,h+i,w+j} \cdot W_{c,h,w,i,j}.
    \label{eq.dpsconv}
\end{equation}
The computational complexity of standard convolution is $\textup{O}(O\times C\times H \times W \times k^2 )$, while DPSConv is $\textup{O}( C\times H \times W \times k^2)$, same as DWConv.

The overall design of RAT is depicted in Fig.~\ref{map}~(b). We expand the prototypes based on the positional information of the masks, filling in the corresponding copies of the prototypes at each pixel position. Then, we \wh{feed-forward} the input through Channel Transform Layers (CTL) and concatenate it with the expanded features. The resulting variable, which contains both semantics and contexts, is used for subsequent filter generation. 
The DPSConv Kernels Generator (DKG) is designed for generating the filters. In the first layer of DKG, we employ a 1x1 convolution to attempt the fusion of the prototype and the pixel's intrinsic attribute. 
We opt for grouped convolution in the final layer of the DKG to reduce the computation and parameters. 
Then Channel Attention Generator (CAG) generates the channel attention to be multiplied with the output of DPSConv along the channel dimension. Finally, the output passes through a 1x1 convolution to merge channel information.
We apply the RAT in both the Encoder and Decoder, as well as the entropy encoding module. 

\noindent \textbf{Entropy Encoding Module.}  
We use the Channel-wise Auto-Regressive Model (ChARM)~\cite{minnen2020channel} as our entropy estimation module, and the specific hyperparameter configuration follows \cite{stf2022}. 
In this model, the latent variable $y$ is partitioned into 10 groups along the channel dimension. The Gaussian prior distribution $p(y_i|\hat{z}, \hat{y}{<i})$ is estimated in an auto-regressive manner, where the mean and scale of $y_i$ depend on the quantized latent in the preceding groups $\hat{y}_{<i}$ (see Eq.~\ref{eq.z}).
To further enhance the estimation ability, we also deploy RAT in it. The RAT here can transmit key region prototypes through a separate bitstream, assisting in more efficient encoding and decoding. 
The prototypes $p$ could be seen as segmentation-prior, a supplement of the hyper-prior information $z$.

\begin{figure*}[t]
  \centering
  \includegraphics[width=\linewidth]{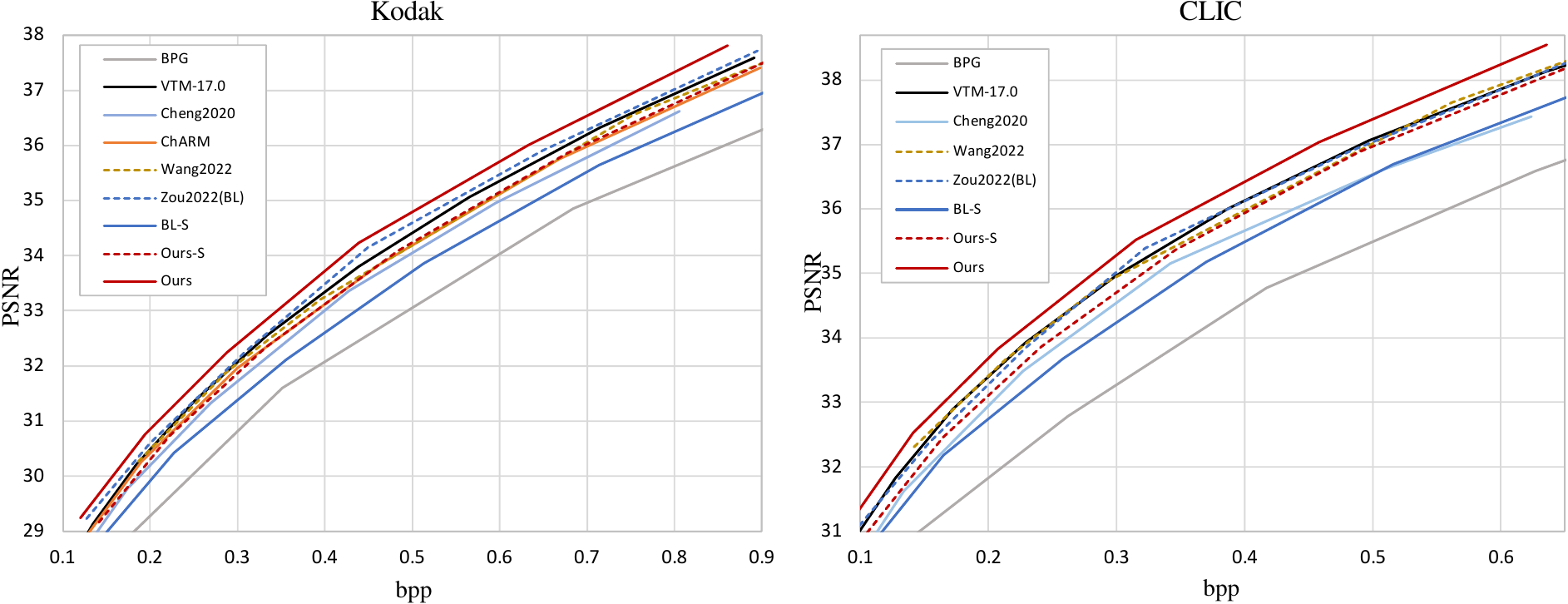}
  \caption{ 
  Rate-Distortion Performance on Kodak~\cite{kodak} and CLIC Professional Valid dataset~\cite{clic}. 
  Our SegPIC introduces proposed RAT and SAL based on Zou2022~\cite{stf2022}. 
  We also implement a simple baseline model without self-attention and auto-regressive mechanism (BL-S), 
  and incorporate our proposed modules on it (Ours-S).
  }
  \label{psnr}
\end{figure*}

\begin{table}[t]
    \caption{\new{BD-rate(\%) performance on Kodak and CLIC Professional Valid (CLIC for short). The anchor is VTM-17.0. }}
    \centering
    \begin{tabular}{lcc|lcc}
        \toprule[1pt]
 & \textit{Kodak}&\textit{CLIC} & & \textit{Kodak}&\textit{CLIC}\\
         Model&  BD-rate↓ &  BD-rate↓ \quad &  \quad Model&  BD-rate↓ &  BD-rate↓ \\
         \midrule
         BPG~\cite{bpg}&  +31.5    &  +48.1     \quad &  \quad EVC~\cite{wang2023EVC}& +0.27    &  --    \\
         Cheng2020~\cite{cheng2020learned}   &  +7.46    & +11.7 \quad &  \quad Feng2023~\cite{feng2023semantically}&   -2.21   &  -- \\
         Agustsson2023~\cite{agustsson2023multi}&  +6.15    & --  \quad     &  \quad ELIC~\cite{he2022elic} &  -3.08&  -3.58  \\
         ChARM~\cite{minnen2020channel} &  +3.58    & -- \quad &  \quad Contextformer~\cite{koyuncu2022contextformer} &  -5.38    &  --\\
         Wang2022~\cite{wang2022neural}      & +1.10    & -0.99 \quad &  \quad  Pan2022~\cite{pan2022content} &  -5.42    &  --    \\
         \midrule
         BL-S&  +15.5    & +16.6 \quad &  \quad Zou2022 (BL)~\cite{stf2022}&  -2.89    &  +0.14   \\
         Ours-S& +3.87    & +5.26 \quad &  \quad \textbf{Ours}&  \textbf{-8.18} & \textbf{-8.26} \\
         \bottomrule[1pt]
    \end{tabular}
    \label{tab:bdbr}
\end{table}

\section{Experiments}
\subsection{Experimental Setup}

\noindent \textbf{Training.} 
We use COCO-Stuff~\cite{coco} as the training dataset, which is composed of 118k images with panoptic semantic masks. We just use the class-agnostic masks without the category labels. 
We randomly crop the images with the size of $256 \times 256$. 
The models are trained for 270 epochs (about 1M iterations) using the Adam optimizer~\cite{kingma2014adam} with a batch size of 32. 
The initial learning rate is set to $1 \times 10^{-4}$ for 170 epochs, and reduced by half every 30 epochs.
Our models are optimized by the rate-distortion trade-off loss function (see Eq.~\ref{eq.loss} and Eq.~\ref{eq.ours}). The $\lambda$ belongs to \{0.0018, 0.0035, 0.0067, 0.0130, 0.0250, 0.0483\} for MSE as the distortion,
and \{2.4, 4.58, 8.73, 16.64, 31.73, 60.50\} for MS-SSIM~\cite{msssim}.

\noindent \textbf{Evaluation.} 
We evaluate our model on two commonly used datasets, Kodak \cite{kodak} with the size of $768 \times 512$, 
and CLIC Professional Validation \cite{clic} with the size of 2K. Performance is evaluated using both bitrate and distortions. 
Distortion measurement involves PSNR and MS-SSIM. And bitrates are assessed in bits per pixel (bpp). In this phase, we replace the masks with $4 \times 4$ grid partitions.
Thus, there is no bitrate cost related to the masks, while the bitrate of prototypes $R(p)$ and $R(p')$ is still taken into account.

\subsection{Rate-Distortion Performance}
We compare our SegPIC with previously well-performing methods, including discretized Gaussian mixture likelihoods and attention modules \cite{cheng2020learned} (denoted as Cheng2020), ChARM~\cite{minnen2020channel} (denoted as Minnen2022), ChARM with WAM~\cite{stf2022} (denoted as Zou2022),  image-specific data-dependent transform model \cite{wang2022neural} (denoted as Wang2022), 
the best-performing traditional method VTM-17.0~\cite{vvc} and influential traditional method BPG~\cite{bpg}. 
Furthermore, we introduce two simplified models without self-attention and auto-regressive mechanism, denoted as baseline-small (BL-S) and the one incorporating our proposed SAL and RAT (Ours-S), to demonstrate the generality of our modules. Details about the architecture of these two models will be provided in the appendix.

As illustrated in Fig.~\ref{psnr} and Tab.~\ref{tab:bdbr}, compared to VTM-17.0, our SegPIC could save 8.18\% BD-rate~\cite{bdbr} on Kodak and 8.26\% on CLIC, which outperforms all the LIC and traditional methods mentioned above.
To further verify the efficiency of the proposed RAT and SAL,
We also compare SegPIC with the baseline Zou2022, which can save 5.45\% BD-rate on Kodak, and 8.50\% on CLIC (see Tab.~\ref{tab:ablation}). 
Moreover, Ours-S even achieves closely matched performance compared to ChARM.
The MS-SSIM results are provided in the appendix.

\subsection{Ablation Study}

\begin{figure*}[t]
  \centering
  \includegraphics[width=\linewidth]{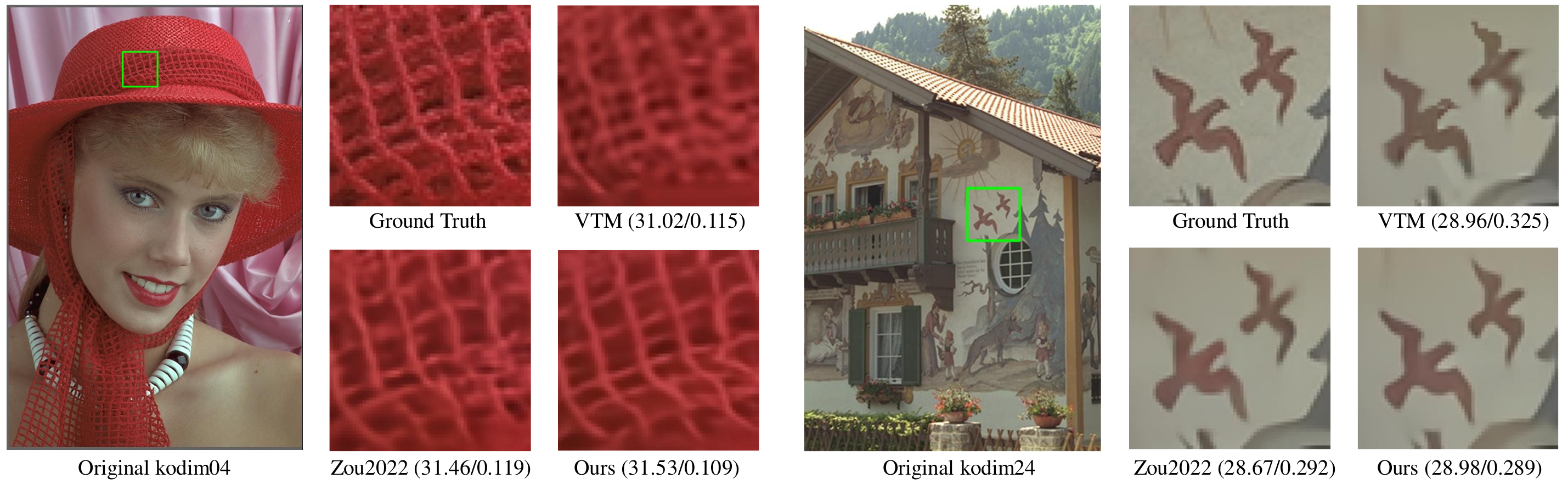}
  \caption{
  Visualization of the reconstructed images \textit{kodim04} and \textit{kodim24} in Kodak. The metrics are (PNSR↑/bpp↓). 
  It shows that our SegPIC can distinguish the objects' contours more accurately, making the edges sharper with less bitrate.
  }
  \label{vis}
\end{figure*}
\begin{table}[t]
    \caption{\new{Ablation Study on Kodak and CLIC. The Baseline~(BL) is Zou2022~\cite{stf2022}. The speed is measured on Kodak with one Nvidia RTX 3090 GPU.}}
    \centering
    \begin{tabular}{lccccc}
    \toprule[1pt]
        & & \textit{Kodak}&\textit{CLIC}\\
         Model          & Params  & BD-rate ↓    & BD-rate ↓ & Enc Time  & Dec Time \\
         \midrule
         Zou2022~(BL)   & 75.2 M   &  0.0    & 0.0 &  120 ms    & 124 ms \\
         BL w/ SAL      & 76.7 M   &  -2.12   & -2.93 &  --  & -- \\   
         BL w/ RAT      & 82.0 M   &  -2.75   & -4.35 &  --  & --  \\   
        Ours           & 83.5 M   &  -5.45  & -8.50  &  142 ms  & 130 ms \\
        \hline
         ChARM     & 48.8 M   &  +6.52    & --  &  107 ms  & 108 ms \\
         BL-S           & 21.7 M   &  +19.0  & +16.3 &  --  & --  \\
         Ours-S         & 29.2 M   &  +6.96  & +5.11  &  99 ms  & 46 ms \\
    \bottomrule[1pt]
    \end{tabular}
    \label{tab:ablation}
\end{table}
To assess the contributions of our RAT and SAL modules individually, we independently incorporate these two modules into the baseline Zou2022. Consequently, we construct two models: ``baseline w/ SAL'' and ``baseline w/ RAT.'' As illustrated in Tab.~\ref{tab:ablation}, the parameter counts introduced by RAT and SAL separately are 1.5M and 5.8M, which occupy only a small fraction of the total parameter count, about 1.7\% and 6.9\%.
And they can contribute to BD-rate saving by 2.93\% and 4.35\% on CLIC Professional Valid, respectively. 
It shows that the proposed RAT and SAL are lightweight and efficient. 
When we combine these two modules, SAL will provide stronger prototypes for RAT, leading to better results, \textit{i.e.}, a saving of 8.50\% BD-rate.

With our proposed modules, Ours-S achieves high performance with a significantly smaller parameter count. Compared to ChARM~\cite{minnen2020channel}, Ours-S saves nearly 40\% parameters and has faster speed while achieving comparable performance. 
Ours-S is a CNN-based model without any self-attention module. And the entropy estimation is just the hyperprior-style \cite{balle2018}, without the efficient but time-consuming auto-regressive architecture. It indicates that the proposed modules RAT and SAL are highly effective and lightweight.

\subsection{Subjective Results}

Fig.~\ref{vis} shows the subjective results compared with VTM and Zou2022. 
As we can see, with the help of the segmentation masks in training, our SegPIC can distinguish the objects' contours more accurately, making the edges sharper with less bitrate. In \textit{kodim04}, we can see the wool reconstructed by SegPIC is more consistent.
In \textit{kodim24}, the bird patterns decoded by SegPIC are more clear, and the blurrings are less than in other methods.
These results are based on grid partitions used as alternatives to masks. 
It demonstrates that SegPIC has acquired a more robust representative capability for various objects. Consequently, SegPIC can discern objects in complex textures more effectively.

\begin{figure*}[t]
  \centering
  \includegraphics[width=\linewidth]{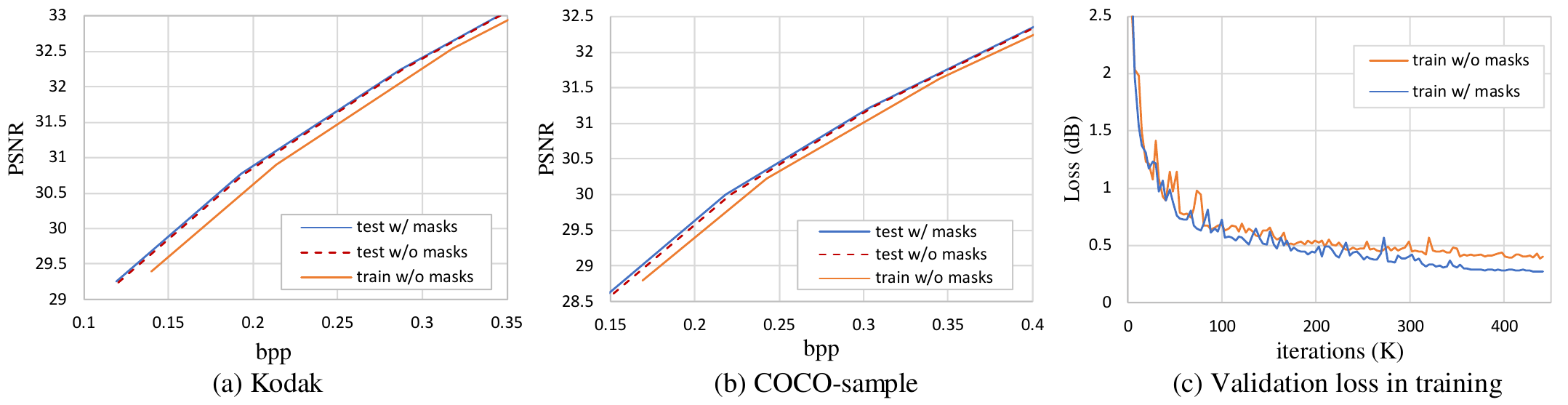}
  \caption{
      We measure the impact of the masks separately in both the training and testing phases. ``test w/ masks'' means we adopt the masks in both training and testing, ``test w/o masks'' means we adopt the masks only in training, while ``train w/o masks'' means we do not use masks all the time. Naturally, if the masks are not used, 4×4 grid partitions will be used as alternatives.
  }
  \label{womasks}
\end{figure*}

\subsection{Analysis of the Role of Segmentation Masks}

\noindent \textbf{Quantitative Results.}
To validate the role of the masks, we make experiments on two datasets, Kodak and COCO-sample. 
Due to Kodak not being a segmentation dataset, the corresponding masks are generated using the segmentation method \cite{entity}. Otherwise, we randomly sample 20 images and their masks from COCO Valid 2017~\cite{coco} to create the COCO-sample dataset. 

As illustrated in Fig.~\ref{womasks}~(a)~(b), during the testing~(inference) phase, the differences caused by masks are not so evident. However, for training, the situation is entirely different. With the assistance of the masks, the model achieves significantly better performance, especially at low bitrates. It could imply that, due to bitrate loss constraints during training, the information in the prototypes is implicitly transferred to the learnable parameters in RAT, allowing limited contextual information to still be effective during the inference phase. At low bitrates, since the lack of sufficient high-frequency information, pixels in the region exhibit stronger similarity, enhancing the representational ability of features extracted by prototypes. 
Fig.~\ref{womasks}~(c) shows the early stage of model training.

\new{
As shown in Tab.~\ref{tab:re2}, the model training w/o masks (the last column) \wht{leads to} noticeable degradation. 
However, training w/ masks has a relatively minor impact on the pattern and number in testing.
\wht{So we can replace the masks in testing to further save bitrate but should take the masks as privileged information during training.}
The hyperparameter 4× in testing is an empirical trade-off between the effect of RAT (\textit{i.e.} image reconstruction precision) and the quantity of the prototypes (\textit{i.e.} the additional bitrate they introduced).
}

\begin{figure*}[t]
  \centering
  \includegraphics[width=\linewidth]{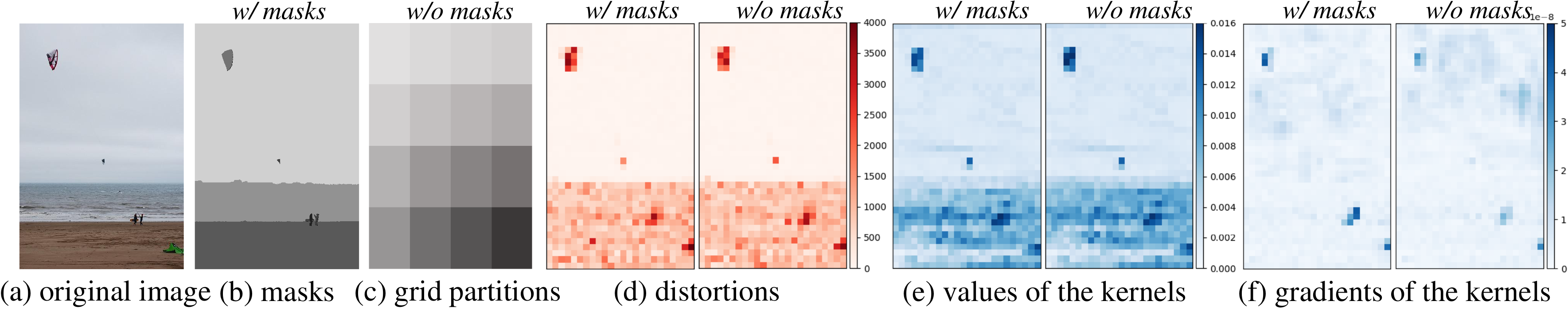}
  \caption{
      Visualization of the kernels generated by DKG in RAT at the Encoder (see Fig.~\ref{map}), in the cases with and without the masks. 
      Fig.~(d)(e)(f) are generated by the same pre-trained network weights. 
      During testing, they go through processes with or without masks, respectively.
      The distortions in (d) can be formulated as $(x-\hat{x})^2$. 
      We take the maximum absolute values of the tensors along the channels and apply 2D max pooling on them for better visualization. 
  }
  \label{gradient}
\end{figure*}

\begin{table}[t]
\caption{\new{The effect of the masks and the grid number on BD-rate on COCO-sample. ``n×'' denotes n×n grid partitions to replace masks.}}
    \centering
    \begin{tabular}{l|cccc|c}
            \toprule[1pt]
   \quad\textbf{Train} \quad & \multicolumn{4}{c|}{train w/ masks}&4×\quad\\
   \hline
           \quad \textbf{Test} \quad & \quad test w/ masks&2×&  4×& 8×&4×\quad\\
          \midrule[1pt]
           \quad \textbf{BD-rate ↓} \quad & \quad -0.11\quad &+0.21 \quad &  0(\textit{default})\quad & +0.46 \quad & +2.10\quad\\
          \bottomrule[1pt]
    \end{tabular}
    
    \label{tab:re2}
\end{table}

\noindent \textbf{Qualitative Analysis.}
Fig.~\ref{gradient} shows the visualization of the kernels generated by DKG in RAT at the Encoder (see Fig.~\ref{arch}). 
Although there is not a noticeable difference in Fig.~\ref{gradient} (e),
it is obvious that the gradients of \textit{w/ mask} are more focused on the semantic objects (see Fig.~\ref{gradient} (f)), which is complex and prone to causing distortions. This attribute will assist in optimizing the models, enabling it to better reconstruct various objects.
These observations indicate that the masks do play a necessary role during the training phase.

\section{Conclusion}
We propose a novel image compression framework called Segmentation-Prior-Guided Image Compression (SegPIC), equipped with two proposed modules: Region-Adaptive Transform (RAT) and Scale Affine Layer (SAL). We adopt the class-agnostic masks as privilege information to assist the training of SegPIC. Experimental results demonstrate that our SegPIC is superior to the previous works. Otherwise, we analyze the role of the masks.
Experiments and visualization results show that the masks can guide the model to focus more on semantic objects in training, leading to higher pixel-fidelity performance.

\section*{Acknowledgements}
This work was supported in part by the National Key R\&D Program of China (No.2021ZD0112100), National NSF of China (No.62120106009),  Guangdong Basic and Applied Basic Research Foundation (2024A1515010454), the Basic and Frontier Research Project of PCL, and the Major Key Project of PCL.

\bibliographystyle{splncs04}
\bibliography{main}

\begin{thebibliography}{10}
\providecommand{\url}[1]{\texttt{#1}}
\providecommand{\urlprefix}{URL }
\providecommand{\doi}[1]{https://doi.org/#1}

\bibitem{clic}
Workshop and challenge on learned image compression (clic2020) (2020), \url{http://www.compression.cc}

\bibitem{agustsson2023multi}
Agustsson, E., Minnen, D., Toderici, G., Mentzer, F.: Multi-realism image compression with a conditional generator. In: IEEE Conf. Comput. Vis. Pattern Recog. pp. 22324--22333 (2023)

\bibitem{dct}
Ahmed, N., Natarajan, T., Rao, K.R.: Discrete cosine transform. IEEE Transactions on Computers  \textbf{100}(1),  90--93 (1974)

\bibitem{akbari2019dsslic}
Akbari, M., Liang, J., Han, J.: Dsslic: Deep semantic segmentation-based layered image compression. In: ICASSP. pp. 2042--2046. IEEE (2019)

\bibitem{balle2016}
Ball{\'e}, J., Laparra, V., Simoncelli, E.P.: End-to-end optimized image compression. In: Int. Conf. Learn. Represent. (2017)

\bibitem{balle2018}
Ball{\'e}, J., Minnen, D., Singh, S., Hwang, S.J., Johnston, N.: Variational image compression with a scale hyperprior. In: Int. Conf. Learn. Represent. (2018)

\bibitem{bpg}
Bellard, F.: Bpg image format (2014), \url{https://bellard.org/bpg/}

\bibitem{bdbr}
Bjontegaard, G.: Calculation of average psnr differences between rd-curves. ITU SG16 Doc. VCEG-M33  (2001)

\bibitem{coco}
Caesar, H., Uijlings, J., Ferrari, V.: {COCO-Stuff}: Thing and stuff classes in context. In: IEEE Conf. Comput. Vis. Pattern Recog. pp. 1209--1218 (2018)

\bibitem{chang19Layered}
Chang, J., Mao, Q., Zhao, Z., Wang, S., Wang, S., Zhu, H., Ma, S.: Layered conceptual image compression via deep semantic synthesis. In: IEEE Int. Conf. Image Process. pp. 694--698 (2019)

\bibitem{chang2021thousand}
Chang, J., Zhao, Z., Yang, L., Jia, C., Zhang, J., Ma, S.: Thousand to one: Semantic prior modeling for conceptual coding. In: Int. Conf. Multimedia and Expo. pp.~1--6. IEEE (2021)

\bibitem{cheng2020learned}
Cheng, Z., Sun, H., Takeuchi, M., Katto, J.: Learned image compression with discretized gaussian mixture likelihoods and attention modules. In: IEEE Conf. Comput. Vis. Pattern Recog. pp. 7939--7948 (2020)

\bibitem{chollet2017xception}
Chollet, F.: Xception: Deep learning with depthwise separable convolutions. In: IEEE Conf. Comput. Vis. Pattern Recog. pp. 1251--1258 (2017)

\bibitem{wavelet}
Daubechies, I.: The wavelet transform, time-frequency localization and signal analysis. IEEE Transactions on Information Theory  \textbf{36}(5),  961--1005 (1990)

\bibitem{vit}
Dosovitskiy, A., Beyer, L., Kolesnikov, A., Weissenborn, D., Zhai, X., Unterthiner, T., Dehghani, M., Minderer, M., Heigold, G., Gelly, S., Uszkoreit, J., Houlsby, N.: An image is worth 16x16 words: Transformers for image recognition at scale. In: Int. Conf. Learn. Represent. (2021)

\bibitem{feng2023semantically}
Feng, R., Gao, Y., Jin, X., Feng, R., Chen, Z.: Semantically structured image compression via irregular group-based decoupling. In: Int. Conf. Comput. Vis. pp. 17237--17247 (2023)

\bibitem{gan}
Goodfellow, I., Pouget-Abadie, J., Mirza, M., Xu, B., Warde-Farley, D., Ozair, S., Courville, A., Bengio, Y.: Generative adversarial nets. In: Ghahramani, Z., Welling, M., Cortes, C., Lawrence, N., Weinberger, K. (eds.) Adv. Neural Inform. Process. Syst. vol.~27. Curran Associates, Inc. (2014)

\bibitem{he2022elic}
He, D., Yang, Z., Peng, W., Ma, R., Qin, H., Wang, Y.: Elic: Efficient learned image compression with unevenly grouped space-channel contextual adaptive coding. In: IEEE Conf. Comput. Vis. Pattern Recog. pp. 5718--5727 (2022)

\bibitem{he2021checkerboard}
He, D., Zheng, Y., Sun, B., Wang, Y., Qin, H.: Checkerboard context model for efficient learned image compression. In: IEEE Conf. Comput. Vis. Pattern Recog. pp. 14771--14780 (2021)

\bibitem{he2016deep}
He, K., Zhang, X., Ren, S., Sun, J.: Deep residual learning for image recognition. In: IEEE Conf. Comput. Vis. Pattern Recog. pp. 770--778 (2016)

\bibitem{hoang2020image}
Hoang, T.M., Zhou, J., Fan, Y.: Image compression with encoder-decoder matched semantic segmentation. In: IEEE Conf. Comput. Vis. Pattern Recog. Worksh. pp. 160--161 (2020)

\bibitem{jia2016dynamic}
Jia, X., De~Brabandere, B., Tuytelaars, T., Gool, L.V.: Dynamic filter networks. Adv. Neural Inform. Process. Syst.  \textbf{29} (2016)

\bibitem{jiang2023mlic}
Jiang, W., Yang, J., Zhai, Y., Ning, P., Gao, F., Wang, R.: Mlic: Multi-reference entropy model for learned image compression. In: ACM Int. Conf. Multimedia. pp. 7618--7627 (2023)

\bibitem{vvc}
(JVET), J.V.E.T.: Versatile video coding (2021), \url{https://jvet.hhi.fraunhofer.de/}

\bibitem{kingma2014adam}
Kingma, D.P., Ba, J.: Adam: A method for stochastic optimization. arXiv preprint arXiv:1412.6980  (2014)

\bibitem{vae}
Kingma, D.P., Welling, M.: Auto-encoding variational bayes. In: Int. Conf. Learn. Represent. (2014)

\bibitem{kodak}
Kodak, E.: Kodak lossless true color image suite (photocd pcd0992) (1993), \url{http://r0k.us/graphics/kodak/}

\bibitem{koyuncu2022contextformer}
Koyuncu, A.B., Gao, H., Boev, A., Gaikov, G., Alshina, E., Steinbach, E.: Contextformer: A transformer with spatio-channel attention for context modeling in learned image compression. In: Eur. Conf. Comput. Vis. pp. 447--463. Springer (2022)

\bibitem{privilege}
Li, F., Zhang, L., Liu, Z., Lei, J., Li, Z.: Multi-frequency representation enhancement with privilege information for video super-resolution. In: Int. Conf. Comput. Vis. pp. 12814--12825 (2023)

\bibitem{liu2020unified}
Liu, J., Lu, G., Hu, Z., Xu, D.: A unified end-to-end framework for efficient deep image compression. arXiv preprint arXiv:2002.03370  (2020)

\bibitem{liu2023learned}
Liu, J., Sun, H., Katto, J.: Learned image compression with mixed transformer-cnn architectures. In: IEEE Conf. Comput. Vis. Pattern Recog. pp. 14388--14397 (2023)

\bibitem{liu2021swin}
Liu, Z., Lin, Y., Cao, Y., Hu, H., Wei, Y., Zhang, Z., Lin, S., Guo, B.: Swin transformer: Hierarchical vision transformer using shifted windows. In: Int. Conf. Comput. Vis. pp. 10012--10022 (2021)

\bibitem{minnen2018joint}
Minnen, D., Ball{\'e}, J., Toderici, G.D.: Joint autoregressive and hierarchical priors for learned image compression. Adv. Neural Inform. Process. Syst.  \textbf{31} (2018)

\bibitem{minnen2020channel}
Minnen, D., Singh, S.: Channel-wise autoregressive entropy models for learned image compression. In: IEEE Int. Conf. Image Process. pp. 3339--3343. IEEE (2020)

\bibitem{pan2022content}
Pan, G., Lu, G., Hu, Z., Xu, D.: Content adaptive latents and decoder for neural image compression. In: Eur. Conf. Comput. Vis. pp. 556--573. Springer (2022)

\bibitem{entity}
Qi, L., Kuen, J., Wang, Y., Gu, J., Zhao, H., Torr, P., Lin, Z., Jia, J.: Open world entity segmentation. IEEE Trans. Pattern Anal. Mach. Intell.  (2022)

\bibitem{shen2023adaptive}
Shen, H., Zhao, Z.Q., Zhang, W.: Adaptive dynamic filtering network for image denoising. In: AAAI. vol.~37, pp. 2227--2235 (2023)

\bibitem{jpeg2000}
Skodras, A., Christopoulos, C., Ebrahimi, T.: The {JPEG} 2000 still image compression standard. IEEE Signal Processing Magazine  \textbf{18}(5),  36--58 (2001)

\bibitem{klt}
Stark, H., Woods, J.W.: Probability, random processes, and estimation theory for engineers. Prentice-Hall, Inc. (1986)

\bibitem{sun2020semantic}
Sun, S., He, T., Chen, Z.: Semantic structured image coding framework for multiple intelligent applications. IEEE Trans. Circuit Syst. Video Technol.  \textbf{31}(9),  3631--3642 (2020)

\bibitem{vaswani2017attention}
Vaswani, A., Shazeer, N., Parmar, N., Uszkoreit, J., Jones, L., Gomez, A.N., Kaiser, {\L}., Polosukhin, I.: Attention is all you need. Adv. Neural Inform. Process. Syst.  \textbf{30} (2017)

\bibitem{jpeg}
Wallace, G.K.: The {JPEG} still picture compression standard. Communications of the ACM  \textbf{34}(4),  30--44 (1991)

\bibitem{wang2022neural}
Wang, D., Yang, W., Hu, Y., Liu, J.: Neural data-dependent transform for learned image compression. In: IEEE Conf. Comput. Vis. Pattern Recog. pp. 17379--17388 (2022)

\bibitem{wang2023EVC}
Wang, G.H., Li, J., Li, B., Lu, Y.: Evc: Towards real-time neural image compression with mask decay. In: Int. Conf. Learn. Represent. (2023)

\bibitem{18sft}
Wang, X., Yu, K., Dong, C., Loy, C.C.: Recovering realistic texture in image super-resolution by deep spatial feature transform. In: IEEE Conf. Comput. Vis. Pattern Recog. pp. 606--615 (2018)

\bibitem{msssim}
Wang, Z., Simoncelli, E.P., Bovik, A.C.: Multiscale structural similarity for image quality assessment. In: The Thrity-Seventh Asilomar Conference on Signals, Systems \& Computers, 2003. vol.~2, pp. 1398--1402. IEEE (2003)

\bibitem{xu2020unified}
Xu, Y.S., Tseng, S.Y.R., Tseng, Y., Kuo, H.K., Tsai, Y.M.: Unified dynamic convolutional network for super-resolution with variational degradations. In: IEEE Conf. Comput. Vis. Pattern Recog. pp. 12496--12505 (2020)

\bibitem{zhu2021transformer}
Zhu, Y., Yang, Y., Cohen, T.: Transformer-based transform coding. In: Int. Conf. Learn. Represent. (2021)

\bibitem{stf2022}
Zou, R., Song, C., Zhang, Z.: The devil is in the details: Window-based attention for image compression. In: IEEE Conf. Comput. Vis. Pattern Recog. pp. 17492--17501 (2022)

\end{thebibliography}

\clearpage
\setcounter{page}{1}
    \noindent {\fontsize{14pt}{22pt}\selectfont \textbf{Appendix}} 


\section{Network Design of Ours-S and BL-S}

\begin{figure*}[h]
  \centering
  \includegraphics[width=\linewidth]{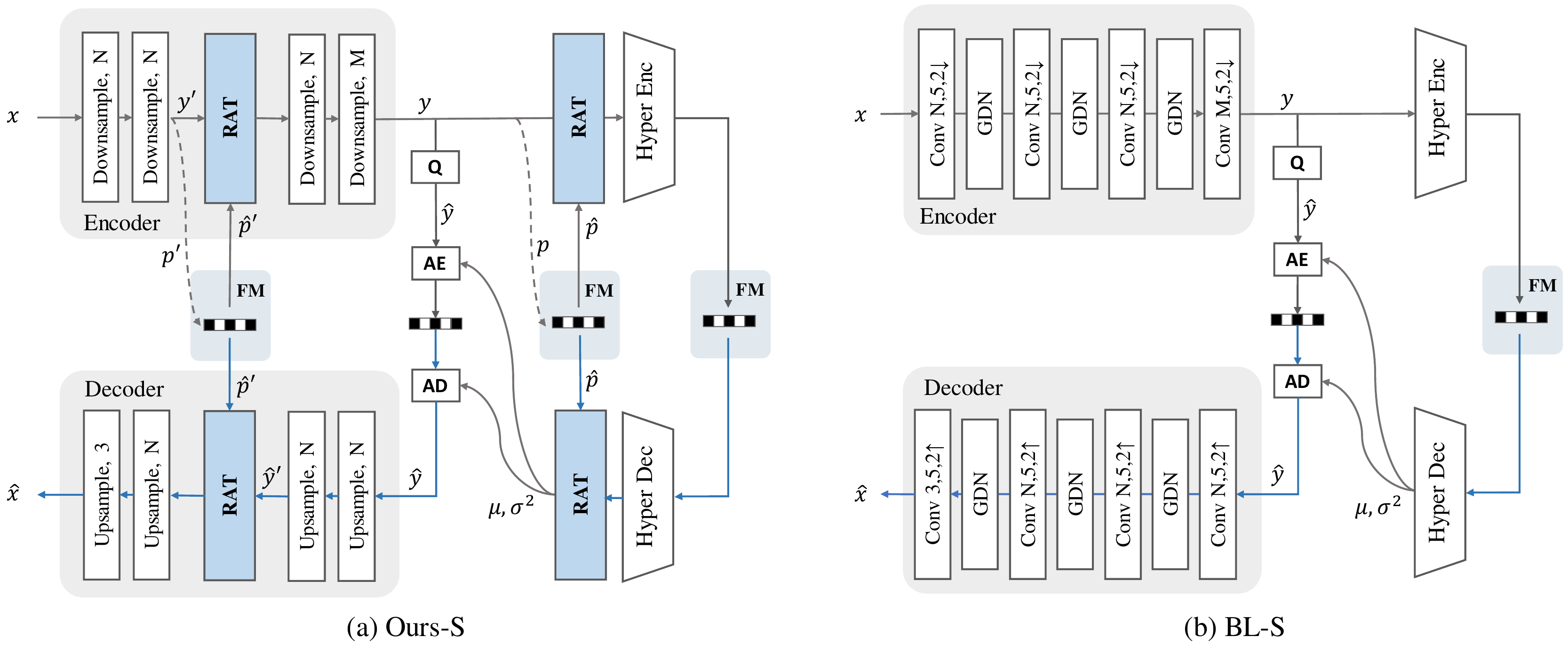}
  \caption{
  The overall framework of Ours-S and BL-S. We construct these two simplified models without self-attention and auto-regressive mechanisms. FM is a Factorized Model~\cite{balle2016}. The details of Hyper Encoder and Decoder are shown in Tab.~\ref{tab:hyper}.}
  \label{ourss}
\end{figure*}

The detailed designs of two simplified models Ours-S and BL-S are illustrated in Fig.~\ref{ourss}. The Hyper Decoder consists of two modules, Hyper Decoder Scales and Hyper Decoder Means, which have identical network structures and each generates the estimated $\sigma$ and $\mu$ for $\hat{y} \sim \mathcal{N}(\mu,\sigma^2)$. The RAT after the Hyper Decoder also consists of both the scale and mean portions, which is not fully depicted in Fig.~\ref{arch}~(a) and Fig.~\ref{ourss} for brevity. The network architectures of Hyper Encoder and Decoder are shown in Tab.~\ref{tab:hyper}.

\section{Rate-Distortion Performance on MS-SSIM}

\begin{figure}[h]
  \centering
  \includegraphics[width=0.6\linewidth]{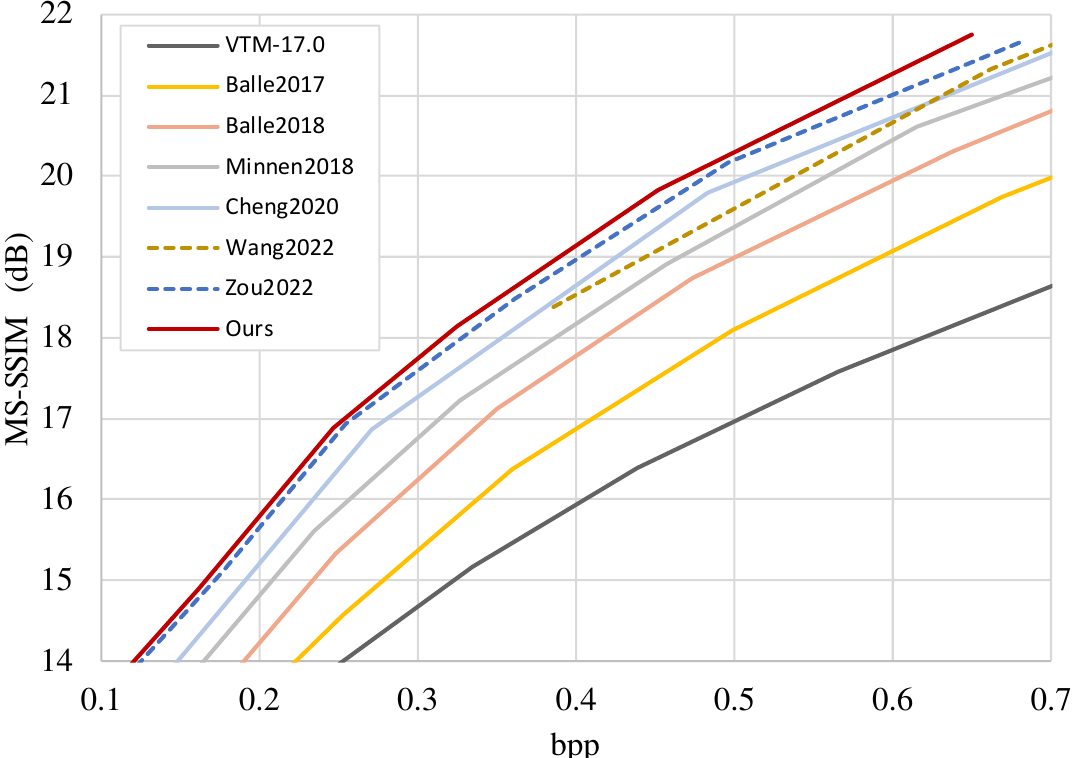}
  \caption{
  The rate-distortion performance on Kodak. The metrics are bpp↓ and MS-SSIM↑. All the LIC methods are optimized by loss function Eq.~\ref{eq.loss}, where the distortions can be formulated as $D = (1-\text{MS-SSIM}(x, \hat{x}))$.
  }
  \label{ssim}
\end{figure}

We also compare the MS-SSIM results with other LIC methods and VTM-17.0. Here we additionally introduce some LIC methods, \textit{e.g.} Factorize Model \cite{balle2016} (denoted as Balle2017), Hyperprior Entropy Model \cite{balle2018} (denoted as Balle2018), Joint Autoregressive and Hierarchical Priors Model \cite{minnen2018joint} (denoted as Minnen2018). 
Fig.~\ref{ssim} shows that our SegPIC outperforms these methods.

\begin{table}[h]
\caption{The network architecture of Hyper Encoder, Hyper Doecoder Scales, and Hyper Decoder Means.
    ``Conv 256,3,2↓" means the convolution operation with out-channel 256, kernel size 3 and downsample step 2. ``TConv" means transposed convolution.}
    \centering
    \begin{tabular}{c|c|c}
    \toprule[1pt]

          Hyper Encoder& Hyper Decoder Scales &  Hyper Decoder Means\\
         \hline
         Conv 320,3,1&  Conv 192,3,1& Conv 192,3,1
\\
         GELU&  GELU& GELU
\\
         Conv 288,3,1&  TConv 224,3,2↑& TConv 224,3,2↑
\\
         GELU&  GELU& GELU
\\
         Conv 256,3,2↓&  Conv 256,3,1& Conv 256,3,1
\\
         GELU&  GELU& GELU
\\
 Conv 224,3,1& TConv 288,3,2↑&TConv 288,3,2↑
\\
 GELU& GELU&GELU
\\
 Conv 192,3,2↓& Conv 320,3,1&Conv 320,3,1\\
 \bottomrule[1pt]
    \end{tabular}
    
    \label{tab:hyper}
\end{table}

\end{document}